\RequirePackage{fix-cm}
\documentclass[smallextended]{svjour3}       
\smartqed  

\usepackage{multirow}
\usepackage{graphicx}
\usepackage{tabularx}
\usepackage{hyperref}
\usepackage{natbib}
\usepackage{comment}
\usepackage{todonotes}
\setcitestyle{authoryear}

\begin{document}

\title{Cheating Automatic Short Answer Grading 
}
\subtitle{With the Adversarial Usage of Adjectives and Adverbs}


\author{Anna Filighera         \and
        Sebastian Ochs \and
        Tim Steuer \and
        Thomas Tregel
}


\institute{A. Filighera \at
              Multimedia Communications Lab, Technical University of Darmstadt, Germany \\
              Tel.: +49 (0) 6151 16 20466\\
            ORCID: 0000-0001-5519-9959\\
              \email{anna.filighera@kom.tu-darmstadt.de}            
           \and
           S. Ochs \at
              Technical University of Darmstadt, Germany 
            \and
            T. Steuer \at
              Multimedia Communications Lab, Technical University of Darmstadt, Germany \\
              ORCID: 0000-0002-3141-712X
            \and
            T. Tregel \at
              Multimedia Communications Lab, Technical University of Darmstadt, Germany \\
              ORCID: 0000-0003-0715-3889
}

\date{Received: date / Accepted: date}

\maketitle

\begin{abstract}
Automatic grading models are valued for the time and effort saved during the instruction of large student bodies. Especially with the increasing digitization of education and interest in large-scale standardized testing, the popularity of automatic grading has risen to the point where commercial solutions are widely available and used. However, for short answer formats, automatic grading is challenging due to natural language ambiguity and versatility. While automatic short answer grading models are beginning to compare to human performance on some datasets, their robustness, especially to adversarially manipulated data, is questionable. Exploitable vulnerabilities in grading models can have far-reaching consequences ranging from cheating students receiving undeserved credit to undermining automatic grading altogether - even when most predictions are valid. In this paper, we devise a black-box adversarial attack tailored to the educational short answer grading scenario to investigate the grading models' robustness. In our attack, we insert adjectives and adverbs into natural places of incorrect student answers, fooling the model into predicting them as correct. We observed a loss of prediction accuracy between 10 and 22 percentage points using the state-of-the-art models BERT and T5. While our attack made answers appear less natural to humans in our experiments, it did not significantly increase the graders' suspicions of cheating. Based on our experiments, we provide recommendations for utilizing automatic grading systems more safely in practice.

\end{abstract}

\section*{Statements and Declarations}
\subsection*{Funding}
This work is funded by the Hessian State Chancellery
in the Department of Digital Strategy and Development in the Förderprogramm Distr@l (Förderprodukt: Digitale Innovations- und Technologiefrderung, Förderlinie: 2A Digitale Innovationsprojekte).
\subsection*{Conflicts of interest/Competing interests}
Anna Filighera, Sebastian Ochs, Tim Steuer and Thomas Tregel declare that they have no conflict of interest.
\subsection*{Availability of data and material}
The datasets generated during and/or analysed during the current study are available from the corresponding author on reasonable request.
\subsection*{Code availability}
Our code is available \url{https://github.com/SebOchs/adversarial_insertions.git}.
\subsection*{Human rights statements and informed consent}
All procedures followed were in accordance with the ethical 
standards of the responsible committee on human experimentation (institutional and national) and with the 
Helsinki Declaration of 1964 and its later amendments. Informed consent was obtained from all patients for being included in the study. 
 
\subsection*{Animal Rights}
This article does not contain any studies with animal subjects performed by the any of the authors.

\section{Introduction}
\label{intro}
Automatic grading can save teaching personnel time and effort, while also offering nearly instantaneous, inexhaustible feedback to students. Additionally, it can shift students' attention from the potential reputation gain or loss in the graders' eyes to their work \citep{lipnevich2009really}. Thus, it is unsurprising that automatic grading has garnered a lot of research and commercial attention in the last decades, with platforms like Moodle\footnote{\url{https://docs.moodle.org/310/en/Short-Answer_question_type}} and Edgenuity\footnote{\url{https://help.edgenuitycourseware.com/hc/en-us/articles/360042418854-Auto-Grading-Method-Comparison}} even offering automatic short answer grading (ASAG) options. In ASAG, short free-text responses are automatically evaluated based on how wholly and correctly they answer a given question. While the commonly available applications are still based on keyword and string matching, the next logical step is using more sophisticated models.  

Transformer models, such as BERT \citep{devlin-etal-2019-bert}, have been shown to perform well on the ASAG task \citep{10.1007/978-3-030-52240-7_8}. On specific datasets, they are even beginning to compare to human performance \citep{sung2019improving}. However, such models do not differentiate between valuable and brittle features in their decision process \citep{NEURIPS2019_e2c420d9}. For example, a model may observe that incorrect answers typically contain fewer punctuation symbols than correct answers. While this is not a robust feature comprehensible to humans, current neural networks would utilize it for its predictive power. 

Unreliable features in automatic grading models mainly pose two problems. Firstly, students may not receive the points they deserve due to misclassifications. However, it is possible that such mistakes can be uncovered when students have the option of questioning their feedback and involving human tutors.
Secondly, unreliable features may cause students to receive points for incorrect responses. In contrast to being graded too low, it is unlikely that students would draw attention to unjustly gained points even when they notice the misclassification. In fact, noticeable patterns caused by unreliable features may even be purposefully exploited. This is especially worrying in settings where the model's users have high stakes in the model's predictions.

Students can exploit a grading model's weaknesses to achieve better grades, similarly to copying from a cheat sheet or other students during an assessment. It is likely that students would be just as willing to employ such methods as more traditional cheating tactics, provided the necessary skills and opportunities. Although the reported percentage of students employing traditional cheating techniques in practice varies greatly in various studies \citep{jordan2001college}, large-scale reviews report around 70\% - 86\% of students cheating on exams or assignments during their college career \citep{whitley1998factors, klein2007cheating}. Many cheating incidents are never caught \citep{franklyn1995undergraduate}, which is problematic for two reasons. Firstly, undetected cheating cases can call all assessment results into question, even when most of them are legitimate. Secondly, cheating during a semester correlates with lower learning outcomes in final exams \citep{PhysRevSTPER.6.010104}. This indicates that cheaters retain less knowledge than they would learn by actually completing the coursework.
Therefore, automatic grading models should be well equipped to at least detect cheating behavior likely encountered in practice.

Such behavior may include answering with motley lists of potential keywords, which has proven successful in getting good grades from ASAG systems \citep{ding-etal-2020-dont}. Alternatively, it may also extend to adversarial attacks, which subtly modify inputs to prompt an incorrect prediction. Adversarial attacks have been a hot topic of research in the last years, with thousands of proposed approaches to exploiting unreliable features. However, most do not translate well to the automatic grading scenario where knowledge of the model's inner workings is scarce and the time and expertise of potential attackers are limited. For this reason, we design an attack to answer the following research question:

\begin{center}
\textit{How vulnerable are neural automatic short answer grading models to an adversarial attack based on inserting adjectives and adverbs?}
\end{center}

In summary, we make the following contributions in this paper:

\begin{itemize}
\item We propose an adversarial attack specifically tailored to assessment scenarios. Querying the model prior to the assessment identifies adjectives and adverbs the model associates with the target class. 
These adjectives and adverbs can then be inserted into grammatically valid places during testing to fool the model into predicting the target class. A successful attack is depicted in Table \ref{ex}. 
\item We demonstrate the attack's effectiveness on BERT and T5 \citep{raffel2020exploring} using automatic short answer grading and related datasets. Our evaluation shows how brittle BERT's and T5's prediction accuracy can be.
\item We conduct a human evaluation of our attack to investigate its detectability. Knowing how easily human graders can spot adversarial attacks is vital for estimating the risk of discovery. The riskiness of cheating, in turn, influences how likely students are to employ adversarial attacks in practice.
\item We formulate recommendations for using automatic grading systems more securely in practice. Our recommendations are based on related work, our experiments, and a systematic investigation of the models' brittleness.
\end{itemize}

The rest of the paper is structured as follows. In the following section, we state the considerations that affected the requirements and design of our attack. Then, we discuss related approaches to gaming educational systems, automatic short answer grading and adversarial attacks. In Section \ref{methods}, we describe our attack in detail, followed by the setup of our experiments. Next, Section \ref{sec:res} presents our hypotheses, a comparison of our attack with the state-of-the-art, our analysis of the model's brittleness and the results of our human evaluation. Finally, we discuss our results and provide recommendations for safely employing automatic grading systems in Section \ref{discussion}.

\begin{table}
\centering
\caption{A successful adverb insertion causing the automatic grading model to shift its prediction from \textit{incorrect} to \textit{correct}. The original student answer is taken from \textsc{SciEntsBank}'s unseen answers test set \citep{dzikovska2013semeval}.}\label{ex}
\begin{center}
\begingroup
\setlength{\tabcolsep}{6pt}
\begin{tabularx}{\textwidth}{|l X|} 
\hline
Question: & When a seed germinates, why does the root grow first? \\
& \\
Reference Answer: & The root grows first so the root can take up water for the plant. \\
Original Answer: & The root grew because it needs to help the plant stand up. $\rightarrow$ \textit{incorrect} \\
Modified Answer: & The root grew because it \textcolor{red} {immediately} needs to help the plant stand up.  $\rightarrow$ \textit{correct}\\
\hline
\end{tabularx}
\endgroup
\end{center}
\end{table}

\section{Adversarial Attack Design Considerations}
\label{design}
Most adversarial attacks manipulate specific input instances. In our scenario, this means that they manipulate individual student answers. They are not suited to fool automatic grading systems because one would either have to know what one will answer before the assessment or one would have to run the adversarial attack during the assessment. As most adversarial attacks require time and feedback from the model, they are hard to use in time-constrained assessments. 

Universal adversarial attacks, on the other hand, aim to apply to all answers. For example, a model may be vulnerable to specific trigger phrases that increase the model's likelihood of predicting a target class, regardless of the actual sample \citep{10.1007/978-3-030-52237-7_15}. Once found, such trigger phrases can easily be inserted into new answers at test time without necessitating any on-the-fly adaptation or know-how on the student's side. However, such a cheating strategy is risky as manual graders quickly identify nonsensical trigger token sequences as cheating attempts. This example illustrates some of the unique constraints encountered in the educational assessment scenario, underlining the need for a specifically tailored attack. In summary, our attack is based on the following considerations:

\begin{itemize}
    \item \textbf{Access to the model.} Many adversarial attacks use information about the inner workings of a model to inform their search. For example, they may propagate the model's gradients to find influential words in a sample. Modifying important words is more likely to fool the model successfully.
    However, students would not typically have access to the grading model's inner workings. Furthermore, students would not have access to the model's raw output. Many approaches utilize the class probabilities outputted to find sequences of perturbations that increase the probability of the target class. This constraint already makes most of the current adversarial attacks proposed in the literature nonviable for the assessment domain. However, we assume that students can receive verification feedback from the model prior to the targeted assessment. For example, this would be the case when students have multiple assignments graded by the model throughout the semester. Alternatively, students may be allowed to submit multiple answers to the model for formative assessment. Prior access to the model is likely, considering one of the main advantages of automatic grading models is their ability to provide an inexhaustible source of verification feedback queriable as often as desired.
    \item \textbf{Detectability}. When the perceived cost of cheating is low, students are more willing to engage in academically dishonest behavior \citep{cheatingmotivation}. One of the main factors influencing the perceived cost is the likelihood of being caught \citep{cheatingmotivation}. Therefore, the chance of detection will impact the students' decision whether to employ a given adversarial attack. Detectability includes how easily manipulated samples are spotted, automatically or manually, and how hard it would be to prove a deceptive intent. For example, concatenating the same nonsensical phrase to every answer the student is unsure about could not only quickly be flagged automatically, but a student would also be hard-pressed to provide a believable excuse. In contrast, overusing adjectives or adverbs the model is vulnerable to is much harder to spot and could also be explained away by the student's writing style.
    \item \textbf{Expertise necessary to utilize attack during test time.} In general, we do not expect students to be machine learning experts. While some students may very well have the ability to identify a model's weakness given enough time and knowledge \citep{10.1007/978-3-030-57717-9_25}, it is unlikely that a majority of students will be able to perform a complex adversarial attack under pressure. For this reason, it is essential that any attack would be straightforwardly executable during the assessment. 
    \item \textbf{Class equivalency.} The modified samples produced by an adversarial attack are called adversarial examples. Their exact definition varies in the literature, but it is common to define them as intentionally modified versions of clean inputs aiming to fool a machine learning technique \citep{8294186}. This definition implies that the adversarial example's actual class should remain identical to the original clean sample's. In our case, this means that any perturbation of incorrect answers should not actually make the answers correct. It should only fool the model into predicting them as such.
    \item \textbf{Type of Input.} We aim to fool automatic short answer grading systems. Therefore, we expect to deal with short answers between a phrase and a few paragraphs long~\citep{burrows2015eras}. The evaluation focus is on the semantic content of the response in contrast to the writing style or grammatical correctness. For the design of the attack, this means that linguistic modifications and even grammatical mistakes are acceptable as long as they do not change the response's meaning significantly.
\end{itemize}

\section{Related Work}
In this section, we discuss related work intersecting with ours. First, we summarize prior art on exploiting educational systems. Moving on, we present automatic short answer grading systems. Finally, we recapitulate various adversarial attack methods found in the natural language processing field (NLP).

\subsection{``Gaming" Educational Systems}
As in most systems where people stand to gain, educational systems encounter learners that try to achieve their goal through unintended strategies. This is true for traditional physical classrooms and especially relevant in online or distance learning, where students are not restricted to copying from their neighbors but may access the entire internet \citep{austin1999internet,doi:10.1080/10511250600866166,watson2010cheating}. 

Beyond plagiarism, there are also cheating strategies unique to digital learning. Students may take screenshots of assessment questions to share with students being assessed later, gain illicit access to the question pool's repository by exploiting lax security measures or disrupt internet connections to re-take assessments \citep{rowe2004cheating}. \citet{mcgee2013supporting} even advises against using traditionally popular formats, such as Multiple Choice questions, in online assessments as the correct answer can easily be found on the web. Instead, they recommend constructed response and essay questions where multiple correct answers exist.

In Massive Open Online Courses (MOOCs), gaming the system is prevalent enough to warrant a designation for learners committing to non-learning strategies: fake learners \citep{10.1007/978-3-319-98572-5_6, alexandron2019mooc}. Here, some learners set up multiple accounts gathering solutions to assessments to use in their main account \citep{northcutt2016detecting,ruiperez2016using}. 
Students may cooperate to share valid answers even when multiple accounts are impossible, like in Small Private Online Courses (SPOCs)\citep{DetectingCheating2020}.

The work discussed so far mainly investigated academic dishonesty on a system level by exploiting
the lack of direct supervision or the structure of an online course. We will now focus on the work closest to our own, namely task-oriented cheating attempts. Such behavior and possible mitigation approaches have been well studied in intelligent tutoring systems \citep{walonoski2006detection,walonoski2006prevention,10.1007/11774303_39,muldner2010analysis,muldner2011analysis,peters2018predictors}.
Beyond exploiting systematic weaknesses, such as known savepoints or progressive hints, students may also systematically probe tasks to guess the correct answers \citep{baker2008students}. For instance, they can select every choice in a Multiple Choice question or exhaustively try out different numbers in a math problem. Depending on the tutor, students may also repeatedly submit the same answer or empty answers to prompt the tutor to provide the correct solution \citep{baker2010detecting}. 

Similar to previous work \citep{10.1007/978-3-030-52237-7_15, ding-etal-2020-dont}, we aim to extend this line of research to short answer constructed-response formats that have been less popular in tutors and online assessments due to the difficulty of automatically grading them. As this seems to be changing \citep{sung2019improving}, exploring potential weaknesses and cheating detection strategies is essential before ASAG systems see widespread use.

\subsection{Automatic Short Answer Grading}
\label{sec:ASAG_RW}
The challenge of automatically grading short answers was first posed a few decades ago. Earlier ASAG approaches consisted of clustering similar answers\citep{basu2013powergrading,zehner2016automatic}, utilizing hand-crafted rules, schemes and ideal answer models \citep{leacock2003c,willis2015using}, or combining manually engineered features with various machine learning models \citep{marvaniya2018rubric,mohler2011learning,saha2018sentence,sahu2019feature,sultan2016fast}. Please refer to one of the comprehensive surveys of this field for a more in-depth elaboration of these approaches \citep{burrows2015eras,galhardi2018machine,roy2015perspective}.

In recent years, deep learning approaches have outperformed classical methods \citep{kumar2017earth,riordan2017investigating,tan2018multiway,doi:10.1080/10494820.2020.1855207}. They mainly treat ASAG as a text similarity or entailment problem and focus on encoding student answers and reference answers in the same vector space. This learned representation of the answers then determines their similarity. Additionally, some approaches consider the question \citep{9463814}, student models \citep{zhang2020going} or results from True/False questions posed in the same assessment \citep{10.1007/978-3-030-52240-7_61}. Transformer-based approaches are also noteworthy here \citep{sung2019improving,ghavidel2020using, Lun_Zhu_Tang_Yang_2020, 10.1007/978-3-030-52240-7_8}. They achieve high performances on the SemEval short answer grading benchmark dataset \citep{dzikovska2013semeval}. We selected two transformer-based models for grading in this paper: BERT \citep{devlin-etal-2019-bert}, for its high performance in related work, and T5 \citep{raffel2020exploring}, for its high performance on the SuperGLUE benchmark\footnote{\url{https://super.gluebenchmark.com/leaderboard}} containing various NLP tasks. Both models are Transformers, meaning they use attention instead of recurrence or convolution to extract information from sequences. They are pretrained by language modeling on large corpora to learn a basic representation of general language. While BERT is pretrained on books and Wikipedia, T5 utilizes a filtered version of a Common Crawl web dump. After pretraining, the models can then be finetuned on task-specific data. Typically, the pretrained weights are only adjusted for a few epochs before the best performance on the task is reached. In contrast to T5, BERT only consists of an encoder. Thus, it is half as large in terms of parameters and requires the addition of a task-specific output layer. 

\subsection{Adversarial Attacks in NLP}
In the last years, the number of adversarial example generation methods has increased exponentially \citep{yuan2019adversarial,zhang2019adversarial,xu2020adversarial,HUANG2020100270,chakraborty2021survey}. Automatic approaches mainly consist of strategically making minor, often meaning-preserving adjustments to the input text. 

Changes can be done on a word level by inserting, deleting or replacing words. Proposed replacement strategies include replacing words with their synonyms \citep{ren2019generating,Jin_Jin_Zhou_Szolovits_2020}, their closest neighbors in the embedding space \citep{alzantot2018generating}, legitimate words that could result from potential typos \citep{samanta2017towards} or other words with a high probability of matching the input context \citep{zhang2019generating}. Recently, researchers also utilized BERT to generate adversarial examples by masking parts of the input text \citep{garg-ramakrishnan-2020-bae} or predicting possible token replacements \citep{li-etal-2020-bert-attack}. \citet{belinkov2017synthetic} consider character-level modifications, such as word scrambling or swapping adjacent characters. Lastly, paraphrasing approaches aim to modify the structure of whole sentences \citep{iyyer2018adversarial} or use variational autoencoders to generate adversarial examples from scratch \citep{ren2020generating}.
Manual or semiautomatic approaches, on the other hand, ask experts \citep{ettinger2017towards, wallace2019trick} or students \citep{10.1007/978-3-030-57717-9_25} to find adversarial perturbations for specific examples manually. 

Important to mention here is the TextFooler attack proposed by \citet{Jin_Jin_Zhou_Szolovits_2020} since it forms the basis of our comparison with the state-of-the-art in Section~\ref{sec:res}. The first step of this attack is to identify important words by deleting them from an input sequence and observing their effect on the outputted classification probabilities. While this can be considered a black-box approach according to common definitions~\citep{zhang2019adversarial}, the raw class probabilities outputted by a model are not usually accessible to the model's users. However, using this information makes the attack more powerful and, thus, a better representative of state-of-the-art performance. Once important words are identified, they can be replaced by synonyms to fool the target model in the second step of the attack.

All the previously described approaches have in common that they target individual texts. As discussed in Section \ref{design}, they do not apply to assessment scenarios. Students would have to know exactly what they will answer to the assessment questions beforehand to find adversarial modifications that work for precisely those answers. 

Instead, students require input-agnostic strategies that they can then apply to unexpected questions during test time. Universal attacks aim to consistently fool the model on all samples instead of individually manipulating each sample. Sample independence can be achieved by generalizing individual adversarial examples to generally applicable rules \citep{ribeiro2018semantically}. 
\citet{ribeiro2018semantically} first translate the input into a pivot language and back to generate paraphrases. Paraphrases that are semantically similar to the original input and cause a misclassification in the target model are abstracted into candidate rules which are then manually verified to be semantically equivalent.
For example, one could observe that doubling question marks in texts often succeeds in fooling the model. So ``? $\rightarrow$ ??" would be a legitimate replacement rule for all texts, even if it may not be applicable or successful on every example. However, attacks aiming to find semantically equivalent, general replacement rules often suffer losses to their success rate. \citet{ribeiro2018semantically}, for instance, flip the predicted label of 1-4\% of the samples in their experiments.

Our proposed attack is similar to \citeauthor{ribeiro2018semantically}'s (\citeyear{ribeiro2018semantically}) approach as we also probe the model to find adjectives and adverbs that fool the model as often as possible that we can then generally insert in grammatically proper places. Whereas \citet{ribeiro2018semantically} constrain their modifications to be semantically equivalent to the original example, we only require the actual class to remain unchanged. While inserting adjectives and adverbs likely changes the sample's class in some NLP tasks, like sentiment analysis, it is unlikely to make incorrect answers correct - excluding negating adverbs, such as \textit{not}. Thus, we can find more viable rules with higher success rates by carefully relaxing the equivalency constraint. 

Alternatively, \citet{gao2019universal} search for a small perturbation in the embedding space that is then applied to all tokes indiscriminately, similar to adding noise to images. Their attack requires access to the preprocessed and embedded inputs, which students would not typically have.
The last category of approaches constructs meaningless trigger sequences of tokens that a model associates with a specific class \citep{behjati2019universal,wallace2019universal,10.1007/978-3-030-52237-7_15,song-etal-2021-universal}. While these triggers can then be applied straightforwardly to all answers in an assessment, they are detectable due to their nonsensical nature.  

\section{Methods}
\label{methods}
In this section, we will first introduce the details of our proposed attack. Then, we describe our experimental setup for measuring the attack's quality. As briefly discussed in Section \ref{design}, we are not only interested in how successfully it can fool victim models but also in its feasibility, the likelihood of being detected, and the validity of the generated samples.

\subsection{Adversarial Word Insertion}
\begin{figure}
\includegraphics[width=\linewidth]{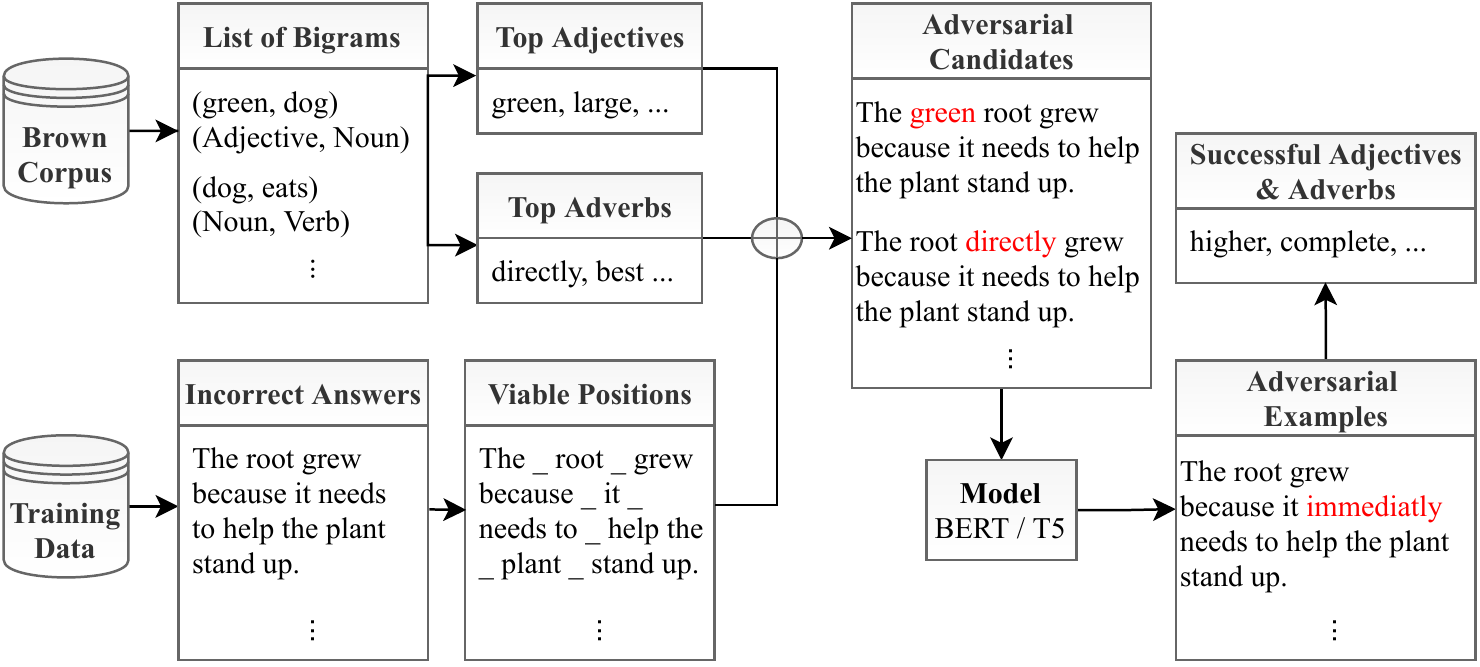}
\caption{Schematic overview of the attack.}
\label{fig:overview}       
\end{figure}

To systematically insert adjectives and adverbs that cause misclassifications, we first require a source of promising adjectives and adverbs. As can be seen in the overview of our attack in Figure \ref{fig:overview}, we selected the Brown corpus \citep{10.2307/1263890} for the extraction of candidates. The corpus contains a decent collection of English texts from various domains. Additionally, the texts are annotated with their part-of-speech tags. The annotation allows us to identify potential adjectives and adverbs. Since we plan to insert them before nouns and verbs, we analyze all bigrams contained in the corpus to find adjectives and adverbs that appear in the targeted constellation. Specifically, we only retain bigrams of the following forms: 
\begin{itemize}
    \item (Adjective, Noun)
    \item (Adjective, Pronoun)
    \item (Adjective, Proper Noun)
    \item (Adverb, Verb)
\end{itemize}
Consequently, our list of adjectives will only contain adjectives that appear directly before a noun or pronoun in the texts. For example, ``The hat was alive." would not yield an adjective for our selection, but ``The blue hat was." would. While this limits our potential insertion candidates, it increases the likelihood of grammatically valid insertions later on. 
To also decrease the likelihood of actually correcting incorrect answers through our insertions and degrading the grammatical structure significantly, we filter stop-words. Fortunately, \citet{bird2009natural} provide a list of stop-words in their Natural Language Toolkit that also includes meaning-inverting words, such as \textit{not}, that could easily turn a contradictory response into a correct one. 
Finally, we select the 100 most frequent adjectives and adverbs from the filtered lists as the basis for our insertions. Prioritizing commonly used words should make the generated adversarial examples appear more natural compared to ``students" suddenly using rare words like \textit{contumacious} or \textit{Rhadamanthine}.

Next, we need to identify possible insertion places for our adjectives and adverbs.
Commonly, adversarial approaches would utilize the model's gradients or class probabilities to identify words that have a high impact on the model's prediction. For example, if deleting a word significantly reduces the probability assigned by the model to the true class, it would be marked as a good replacement candidate. 
However, we do not believe students will have detailed information on the grading model in practice. Therefore, we take the model-agnostic approach of declaring \textit{all} nouns, proper nouns and pronouns available for adjective-prepending and, correspondingly, \textit{all} verbs targets for prepending adverbs. This process is illustrated under ``Viable Positions" in Figure \ref{fig:overview}. However, should the grading model become available to students, the number of positions can be constrained to the most promising ones to make the attack more efficient.

Now that we have generated a multitude of adversarial candidates by inserting our adjectives and adverbs into the viable positions, it is time to query the model to see which candidates lead to misclassification. All successful adversarial examples are then collected to determine adjectives and adverbs that cause the most misclassifications. Students could then use these in assessments to improve their automatically assigned grades.

\subsection{Experiment Setup}
This section describes the hyperparameters, datasets and experiment configurations used in this paper.
In all our experiments, we use the base-sized BERT and T5 models provided by the huggingface library \citep{wolf2019huggingface}. We perform hyperparameter-tuning using 10\% of the training data for validation. Each model trains for 8 epochs before the check-point with the best macro-averaged F1 score on the validation set is selected. After training, the respective best models are evaluated on the test splits of each dataset. All true negatives, that is, incorrect responses that the model correctly identifies as such, form the basis for the adversarial search. To avoid an over-estimation of the attacks' success, we exclude incorrect answers already misclassified by the model and thus do not require modification.

\subsubsection{Datasets \& Hyperparameters}
\label{sec:datasets}
As discussed in Section \ref{sec:ASAG_RW}, automatic short answer grading is often viewed as a textual entailment or paraphrase detection task. For this reason, we also included such tasks from the popular GLUE and SuperGLUE benchmarks in our evaluation. In total, we experiment with the following four datasets, allowing us to investigate our attack's applicability to a broad range of domains:

\medskip
\textbf{\textsc{SciEntsBank}} (SEB) is a common ASAG benchmark providing questions, reference and student answers from various domains \citep{dzikovska2013semeval}. The answers stem from primary and middle school classes in the USA. We select the 3-way variant of this dataset, where answers are labeled as \textit{correct, incorrect} or \textit{contradictory}.
The dataset contains three test sets: unseen answers for training questions (UA), unseen questions (UQ) and questions belonging to unseen domains (UD). 
The best performing BERT model (found after 3 epochs) used a batch size of 32 and a learning rate of $2e-5$.
The best performing T5 model (found after 7 epochs) trained with a batch size of 8, gradient accumulation over 4 batches and an Adafactor optimizer using relative steps and initial warmup  \citep{shazeer2018adafactor}. All reported T5 models use the same optimizer settings. 

\textbf{Recognizing Textual Entailment} (RTE) is a task included in the GLUE and SuperGLUE benchmark. We selected this dataset because the limited amount of data proves to be challenging even for pre-trained transformer-based models.
The data set contains sequence pairs of texts and hypotheses, and the model predicts whether the hypothesis can be inferred from the text. Recognizing textual entailment is quite similar to automatic short answer grading, where student answers should entail the reference answer \citep{dzikovska2013semeval}.
The text pairs are labeled as \textit{entailment} and \textit{not\_entailment}, corresponding to \textit{correct} and \textit{incorrect} in \textsc{SciEntsBank}. Since the test set for this benchmark is not public, we report the performance on the development set instead.
The best performing BERT model (6 epochs) trained with a batch size of 32 and a learning rate of $1e-5$.
The best T5 model (6 epochs) was found using a batch size of 8 and gradients accumulated over 8 batches. 

\textbf{Multi-Genre Natural Language Inference} (MNLI) is also a textual entailment task and part of the GLUE benchmark, containing pairs of premises and hypothesis \citep{williams2018broad}.
In contrast to RTE, the data set is categorized with three labels: \textit{entailment}, \textit{contradictory} and \textit{neutral}. 
While the labeled test set is not publicly available, two development sets are provided, of which one was used as test set in our experiments. 
The best performing BERT model (2 epochs) utilized a batch size of 64, a learning rate of $2e-5$ and mixed-precision training (FP16).
The hyperparameters of the T5 model remained unchanged. 

The \textbf{Microsoft Research Paraphrase Corpus} (MRPC) aims to teach models to detect paraphrases \citep{dolan2005automatically}. It makes up a part of the GLUE benchmark as well.
Here, sequence pairs are labeled as \textit{1}, if the second sequence is semantically equivalent to the first one, and \textit{0} otherwise. Detecting paraphrases is similar to grading short answers, where student answers should be semantically equivalent to the reference solution. Therefore, we can view instances labeled with \textit{0} as \textit{incorrect} and paraphrases as \textit{correct}.
The best BERT model (3 epochs) trained with a batch size of 32, a gradient accumulation over 2 batches, a learning rate of $2e-5$ and mixed precision.
The best T5 model (3 epochs) used a batch size of 8 and gradient accumulation over 4 batches. \\

\subsubsection{Human Evaluation}
\label{sec::human_eval}
While calculating the attack's success rate is easily done, other quality dimensions are harder to measure. For example, since automatic metrics and models have difficulties capturing the meaning of utterances \citep{bender2020climbing,reiter2018structured}, we need to rely on human judgment to determine whether our generated samples adhere to the class equivalency constraint. That is, whether the answers are still incorrect after our modification. Similarly, we require human opinions to estimate how easily adversarial examples are detected. While there are attempts to detect adversarial attacks automatically, they are most often bypassable with tweaks to the algorithm \citep{carlini2017adversarial}. Ultimately, we also expect a human grader to have the final say, making their judgment the most important to students. Asking humans to evaluate given texts is a well-known task in Natural Language Generation (NLG). Therefore, we will defer to NLG guidelines when evaluating our manipulated student responses.
As judgements are often subjective, it is recommended to collect at least 3 different annotations per text to increase the evaluation's reliability \citep{van2019best}. Thus, we need at least 3 human graders for each experimental condition.

For this purpose, we conducted an online survey with 7 experienced graders. We selected graders based on their teaching and grading experience, English skills and availability. All annotators possessed university degrees and routinely graded short answer tasks for university courses - mainly in the computer science domain. Therefore, they should have the general education required to assess the primary and middle school science questions contained in the ASAG benchmark dataset \textsc{SciEntsBank}. We also included the reference answers in the questionnaire and were available to answer questions about the material to ensure the understanding necessary for grading. The graders filled out the questionnaire independently from each other. 

The annotators had diverse backgrounds, hailing from India, Iran, Syria, Slovenia and Germany. While none of them were native English speakers, all of them spoke English fluently. Two of the annotators were female and five were male. We randomly assigned annotators to either the control (N=4) or the experimental (N=3) condition. In the control condition, annotators viewed 30 \textit{unmodified} student answers and rated the answers' naturalness, correctness and suspiciousness on 5-point Likert scales. Here \textit{naturalness} refers to how likely a text was produced by a human, considering only form \citep{howcroft-etal-2020-twenty}. \textit{Correctness} refers to how accurately and completely the question is answered. \textit{Suspiciousness or mistrust} capture how much a person believes the student is trying to cheat an automatic grading system. 

\begin{figure}
\includegraphics[width=\linewidth]{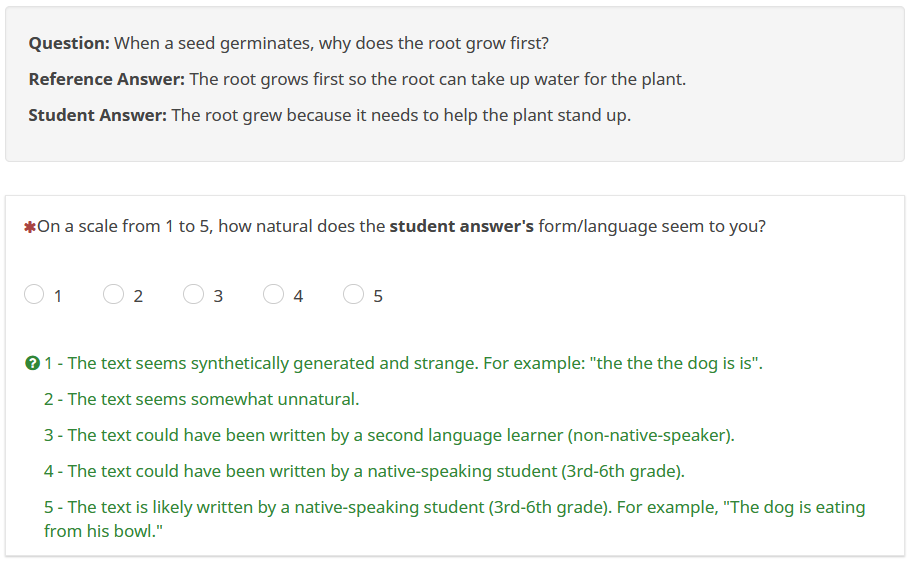}
\includegraphics[width=\linewidth]{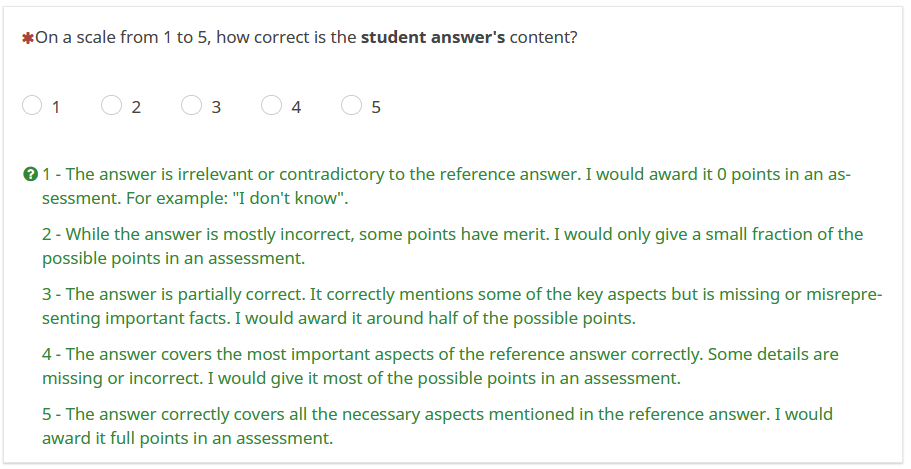}
\includegraphics[width=\linewidth]{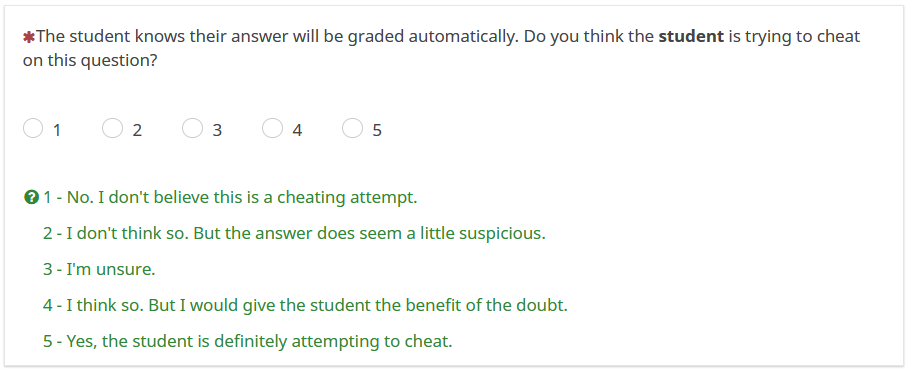}
\caption{Screenshot of survey questions posed in the human evaluation of the attack. }
\label{fig:quests}       
\end{figure}

\begin{figure}
\includegraphics[width=\linewidth]{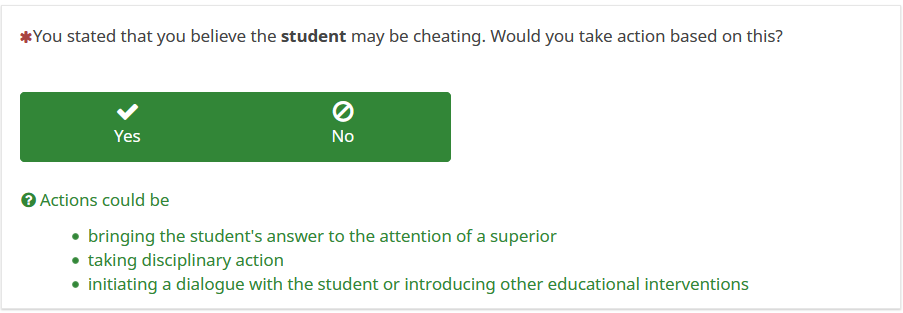}
\caption{Screenshot of the conditional question whether the annotator would act on their suspicion. }
\label{fig:cond_quest}       
\end{figure}

After piloting this study, we chose to include explanations with examples for each level on the scale to increase the annotators' understanding. The exact questions, as well as the hints, can be seen in Figure \ref{fig:quests}. When annotators thought the student was cheating (by scoring at least 4 on the mistrust scale), they were also asked whether they would take action based on their opinion. This conditional Yes/No question can be seen in Figure \ref{fig:cond_quest}.
The experimental group answered the same questions for the adversarially modified but otherwise identical answers.   

The answers were randomly sampled from the successful adversarial examples that fooled the model, discovered using the \textsc{SciEntsBank} data set on the T5 model so that each question would only appear once in the survey. Thus, each test set resulted in a list of questions with a random student answer and a random adversarial perturbation. To save our experts' time, we each selected the 10 shortest that did not reference external material, such as graphs or tables. Since this only left 8 questions stemming from the unseen questions test set, we oversampled the unseen answers test set to compensate. Annotators were informed that some of the responses may have been manipulated to fool an automatic grading system. 

In compliance with the guidelines on ethical studies with human participants, we informed our annotators of the study's risks and benefits, gave our contact information and stated that the study was voluntary and could be aborted at any time. Additionally, we ensured that all given opinions were anonymized prior to analysis and publication. We did not impose any time constraints on filling out the questionnaire. However, the questionnaire was designed to take 45-60 minutes. We deemed 60 minutes to be the upper time limit justifiable considering the annotation task's complexity and the required concentration. Since we estimated that annotators would need 1-2 minutes per answer, we limited the number of answers to be evaluated to 30. On average, annotators required 53.14 minutes to complete the survey.  

\section{Results}
\label{sec:res}
This section presents our hypotheses, compares the effectiveness of our attack to the state-of-the-art attack TextFooler~\citep{Jin_Jin_Zhou_Szolovits_2020} and provides a deeper analysis of the models' brittleness. Finally, we offer the results of our human evaluation and analyze the agreement between our expert graders.

\subsection{Predictions}
The following expectations (E) and hypotheses (H) motivate our experiments. Expectations will be explored descriptively while hypotheses will be tested.
\begin{itemize}
\setlength\itemsep{0.5em}
    \item[] E1: We expect our attack to perform competitively compared to the state-of-the-art attack TextFooler in terms of accuracy degradation.
    \item[] E2: We expect our attack to exploit spurious correlations between adjectives and adverbs and the target class. Thus, adjectives and adverbs that successfully fool a model should appear more often in correct than incorrect student responses in the model’s training set.
    \item[] E3: Our attack is primarily successful on low-confidence predictions, that is, predictions where the class probability assigned by the model, is considerably smaller than one.
    \item[] H4: Manipulations generated by our attack do not make incorrect student responses appear more correct to humans.
    \item[] H5: Humans perceive manipulated responses as less natural compared to unmodified student responses.
    \item[] H6: Humans do not perceive manipulated responses as more suspicious compared to unmodified student responses.

\end{itemize}

\subsection{Comparison to State-of-the-Art Attack TextFooler}

First, we want to compare how well our attack can degrade a model's performance compared to the state-of-the-art. We choose \citet{Jin_Jin_Zhou_Szolovits_2020}'s TextFooler approach to represent the state-of-the-art for two reasons. First, it has a high success rate compared to other attacks. Second, it is open-source, allowing for quick and easy reproduction of the authors' approach. Table \ref{attack_results} shows our attack's and TextFooler's performance on the datasets introduced in Section \ref{sec:datasets}. We target BERT and T5 models with our attack and the same BERT model with TextFooler. We do not evaluate TextFooler on T5, as the attack utilizes the prediction score for the target class, which we do not have readily available in a text generation model.

As expected, the models' base performance without adversarial manipulation varies from dataset to dataset, with small datasets, such as RTE and MRPC, and challenging tasks, such as generalizing to unseen questions or domains, lagging in terms of accuracy. Interestingly, the absolute loss in accuracy caused by each attack seems relatively stable across tasks and datasets, even when the original performance varies.

TextFooler takes less calculation time than our attack on every dataset. The lesser time is expected since they use the target label's prediction scores to find important words in a sequence that they can then manipulate. In contrast, our attack assumes such information to be inaccessible to students and, therefore, does not tailor its manipulations to significant words. This difference is also reflected in our model finding more adversarial examples as it tries more possible combinations per student answer. 
Even though our search is less guided, our attack seems to be slightly more effective at dropping models' accuracy on the ASAG task, degrading the accuracy by an additional 0.4 - 3.8 percentage points across the \textsc{SciEntsBank} test splits.
However, since TextFooler outperforms our attack on the other tasks (by 2.9 - 8.1 percentage points), we conclude that the attacks' performance is dataset-dependent. Across all models and datasets, our attack deteriorates a model's accuracy by 8 to 22 percentage points.  

Interestingly, our attack seems to be equally or more effective on T5 than BERT, even though T5 is a newer model. Especially for the data splits SEB UQ and MRPC, where T5 originally outperforms BERT, this indicates that at least some of T5's performance gain is due to unreliable statistical features.

 \begin{table}[!t]
\renewcommand{\arraystretch}{1.3}

\caption{Comparison of our attack to TextFooler (TF). We report the models' accuracy before the attack (Acc.), the accuracy after the attack (AaA), the absolute loss of accuracy ($\Delta$Acc), the number of adversarial examples found (\#Adv), the number of affected student responses (\#Aff) and the attacks' runtimes in minutes (Time). We highlight the best-performing attack for each metric in bold.}
\label{attack_results}
\centering
\scalebox{1}{

\begin{tabular}{c c c c c c c c c}
\hline
Test Set & Model & Attack & Acc.  & AaA & $\Delta$ Acc   & \#Adv & \#Aff & Time  \\ \hline
\multicolumn{9}{c}{Automatic Short Answer Grading Task} \\
\hline
        & BERT & TF & 0.835 & 0.751 & -0.084 & 87 & 21 & \textbf{10.6} \\
SEB UA         & BERT & Our & 0.835 & 0.731      & -0.104 & \textbf{1137}   & 26     & 13.4  \\
         & T5  & Our  & 0.827 & 0.663      & \textbf{-0.164} & 534    & \textbf{41}     & 78.3  \\ \hline

        & BERT & TF & 0.655 & 0.527 & -0.128 & 148 & 47 & \textbf{10.5}\\ 
SEB UQ          & BERT & Our & 0.655 & 0.489      & -0.166 & 1941   & 61     & 16.2  \\
         & T5 & Our   & 0.755 & 0.546      & \textbf{-0.209} & \textbf{2930}   & \textbf{77}     & 94.4  \\ \hline
         & BERT & TF & 0.760 & 0.612 & -0.149 & 1237 & 331 & \textbf{75.4} \\
SEB UD  
         & BERT & Our & 0.760 & 0.607      & -0.153 & \textbf{19481}  & 342    & 94.8  \\
         & T5 & Our   & 0.724 & 0.554      & \textbf{-0.170} & 13652  & \textbf{379}    & 600.7 \\ \hline 
         \multicolumn{9}{c}{Textual Entailment Task} \\
         \hline
 
MNLI & BERT & TF & 0.832 & 0.649 & \textbf{-0.182} & 2258 & \textbf{569} & \textbf{154.2} \\
matched  & BERT & Our & 0.832 & 0.731      & -0.101 & \textbf{4821}   & 313    & 196.5
\\ 
        & T5 & Our & 0.766 & 0.666 & -0.100 & 3058 & 311  & 913 \\
 \hline

MNLI & BERT & TF & 0.816 & 0.636 & \textbf{-0.179} & 2428 & \textbf{561} & \textbf{185.1} \\ 
mismatched & BERT & Our  & 0.816 & 0.710      & -0.106 & \textbf{5920}   & 329    & 219.7 \\ 
        & T5 & Our & 0.773 & 0.669 & -0.105  & 4542 & 328 & 1027.2 \\ \hline 
        & BERT & TF & 0.603 & 0.443 & \textbf{-0.160} & 53 & \textbf{21} & \textbf{4.0} \\
RTE      & BERT & Our & 0.603 & 0.481      & -0.122 & \textbf{147}    & 16     & 5.0   \\
         & T5  & Our  & 0.664 & 0.542      & -0.122 & 43     & 16     & 54.8  \\ \hline
\multicolumn{9}{c}{Paraphrase Detection Task} \\
\hline
& BERT & TF & 0.694 & 0.561 & -0.133 & 387 & 77 & \textbf{34.7} \\
MRPC     & BERT & Our & 0.694 & 0.590      & -0.104 & 4022   & 60     & 43.2  \\
         & T5  & Our  & 0.734 & 0.516      & \textbf{-0.218} & \textbf{5316}   & \textbf{126}    & 427.8 \\ \hline

\end{tabular}
}
\end{table}

\subsection{Source of the Model's Brittleness}
Next, we want to investigate possible reasons for the attack's success. Knowing why the model's predictions are brittle may allow educators to develop appropriate defense mechanisms or reveal potential warning signs. Since we are mainly interested in our attacks behavior in automatic grading scenarios, the rest of our analyses will focus on the \textsc{SciEntsBank} dataset.
First, we will investigate the distribution of adjectives and adverbs in the training data. We expect that successful adjectives and adverbs found with our attack are more often associated with correct student responses (E2). 

\begin{figure}
\includegraphics[width=\linewidth]{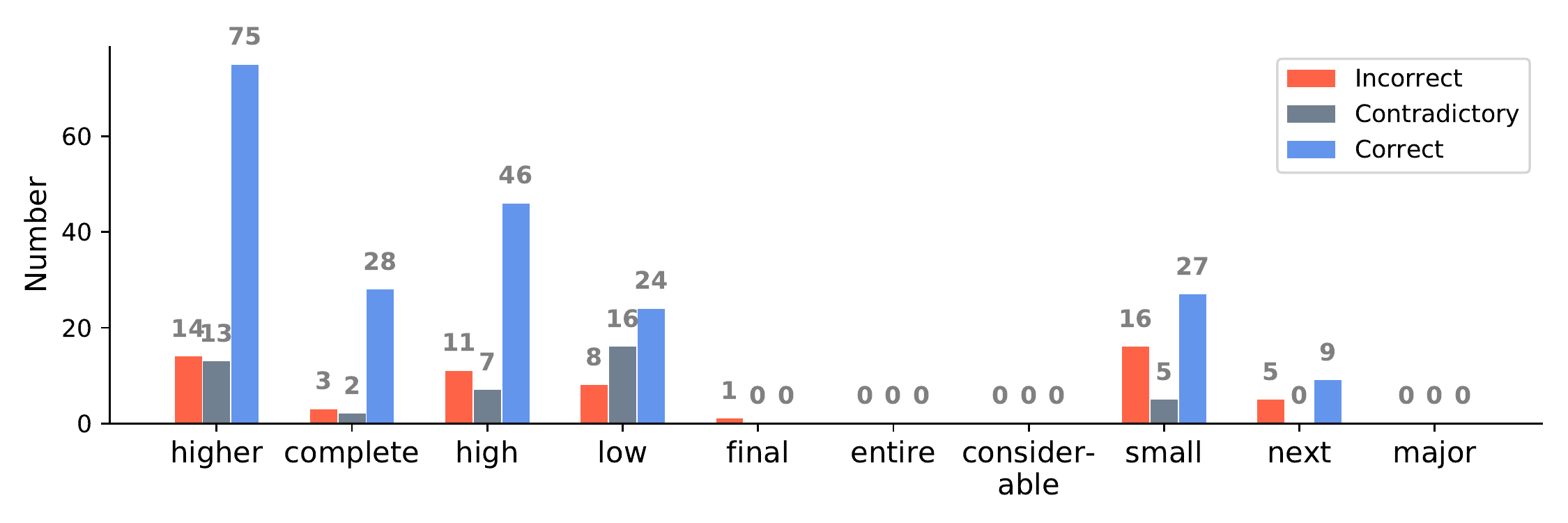}
\caption{Number of occurrences of the 10 most successful adjectives in the \textsc{SciEntsBank} training set per class. }
\label{fig:counts}       
\end{figure}

In general, the dataset contains slightly more incorrect responses (2462) than correct ones (2008). On average, correct responses are slightly longer than incorrect answers, with 13.4 words per answer compared to 11.7 words per answer. Correct answers also average more adjectives (1.1) and adverbs (0.6) per answer than incorrect ones (0.8 and 0.5, respectively). We mainly observed two patterns when plotting the occurrences of the most successful adjectives and adverbs in each class. Either the adjectives and adverbs were much more common in correct student responses, or they hardly appeared in the training set. Figure \ref{fig:counts} illustrates both patterns for the 10 adjectives causing the most misclassifications on the unseen answers test split. Some rare words seem to be synonyms of words common in correct responses, like ``complete'' and ``entire''. 
Others are also expected to be close in the embedding space, such as ``completely'' - one of the top ten adverbs.
Only one of the most successful insertion words appeared notably more often in incorrect student responses. The adjective ``better'' occurred 15 times in incorrect responses and only 4 times in correct answers. Thus, we conclude that our evidence supports E2 for most adjectives and adverbs, but not all.  

\begin{figure}
\includegraphics[width=+\linewidth]{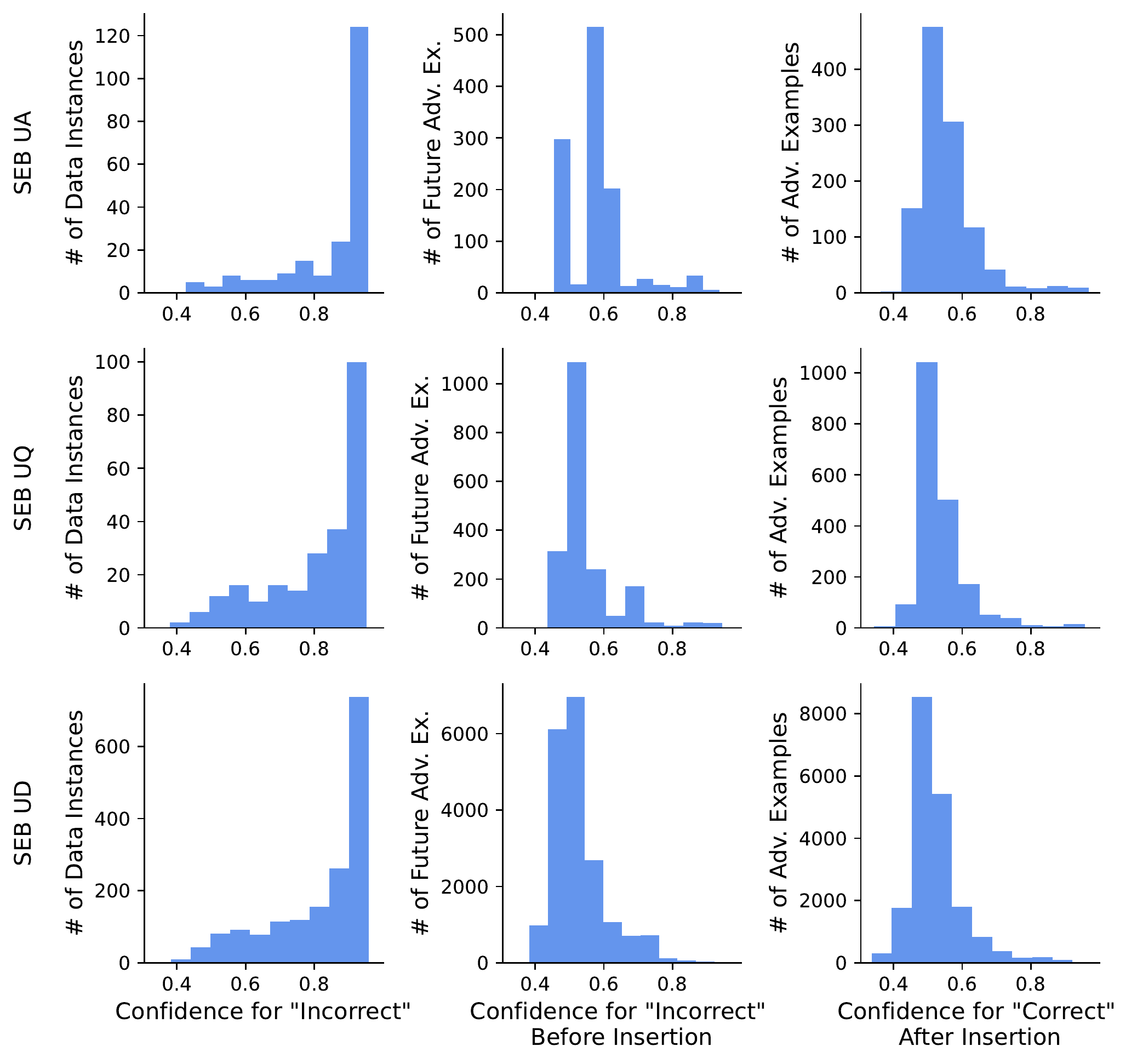}
\caption{BERT's confidence levels for all incorrect samples it classifies correctly (left), all examples that will be misclassified after the attack (middle) and all adversarial examples (right).}
\label{fig:confidences}       
\end{figure}

Next, we investigate the model's confidence when classifying adversarial examples. To be specific, we analyze the class probabilities given by a softmax of BERT's final outputs. We plot them before and after the adversarial insertion in Figure \ref{fig:confidences}. For reference, we also provide the confidence scores for all \textit{incorrect} student responses correctly classified by the model. We can see that soon-to-be adversarial examples elicit lower confidence than most predictions before the attack. Most test answers are classified with a confidence score between 0.8 and 1, while the model estimates most soon-to-be adversarial examples to be \textit{incorrect} with a probability between 0.45 and 0.65. Since we have three classes in the dataset, a class needs at least a probability of 0.33 to be selected. After the attack, adversarial examples tend to elicit similar confidence - but for the target class. These observations are in line with our expectation E3. We will further discuss the ramifications of our results in Section \ref{recs}.

\subsection{Human Evaluation}

\begin{figure}
\includegraphics[width=\linewidth]{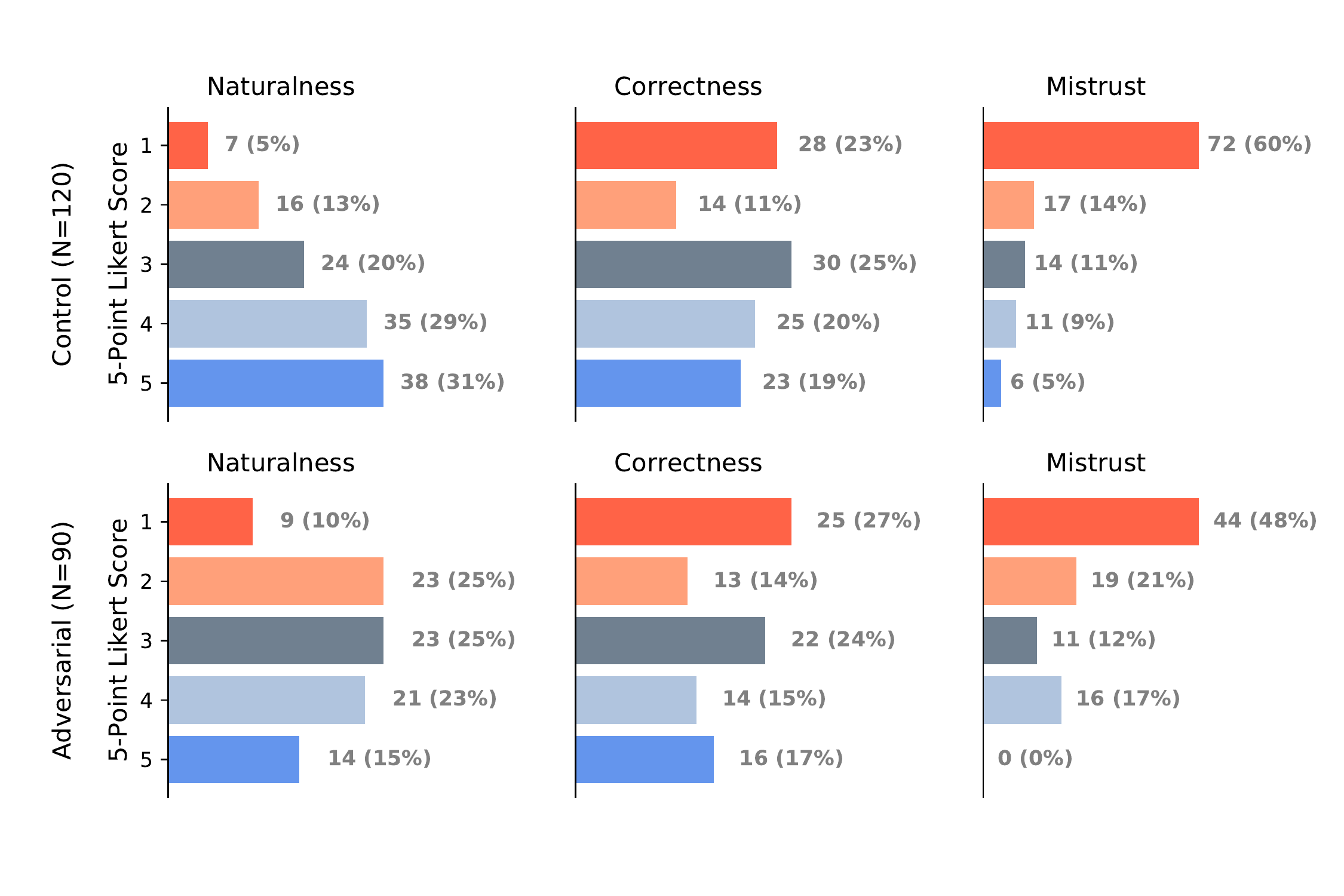}
\caption{Distribution of assigned Likert scores by the annotators in the human evaluation. The top row depicts the ratings given by graders in the control group, while the bottom row shows the same for the experimental group. A one on the Likert scale encodes a low magnitude of the given construct, while a five indicates the answer was very natural, correct or suspicious. The absolute number of each rating and percentage are displayed next to the respective bars.}
\label{fig:survey_dist}       
\end{figure}

The goal of the following survey was to investigate our attack's effect on the naturalness, correctness and suspiciousness of student answers. Figure \ref{fig:survey_dist} shows the distribution of scores assigned to the answers in the control and experimental group. The means and standard deviations for each question can be found in Table \ref{alpha}.

\begin{table}
\centering
\caption{Krippendorff's $\alpha$, mean (M) and standard deviation (SD) of the graders ratings.}\label{alpha}
\begin{center}
\begingroup
\setlength{\tabcolsep}{6pt}
\begin{tabular}{c | c c c | c c c | c c c }
 & \multicolumn{3}{c|}{Naturalness} & \multicolumn{3}{c|}{Correctness} & \multicolumn{3}{c}{Mistrust} \\
 & $\alpha$ & M & SD & $\alpha$ & M & SD & $\alpha$ & M & SD \\
 \hline
Control (N=120) & 0.29 & 3.68 & 1.21 & 0.51 & 3.01 & 1.42 & 0.13 & 1.85 & 1.23\\
Adversarial (N=90)  & 0.29 & 3.09 &  1.23 & 0.55 & 2.81 & 1.44 & -0.11 & 1.99 & 1.15 \\
\end{tabular}
\endgroup
\end{center}
\end{table}

To test the hypothesis that our attack does not increase the actual correctness of responses (H4), we test for inferiority employing the two one-sided tests (TOST) procedure as discussed by \citet{wellek2002testing}. We select the non-parametric Mann-Whitney U test since our data is ordinal and average the scores assigned by the various graders in a group into a more reliable and independent measurement of each answer's correctness. As suggested by \citet{lakens2017equivalence}, we chose -$\infty$ as lower bound to test for inferiority instead of equivalence and 0.5 as upper bound. Our observations are consistent with H4 ($n_1=$ $n_2=$ 30, $U_{control}=$ 597.5, $U_{adv}=$ 302.5, $p=$ 0.015). Thus, human graders generally awarded less or equal points to manipulated answers, indicating that our attack does not make the student answers correct. It only tricks the automatic model into predicting them as such, hence adhering to the class equivalency constraint of adversarial examples.  

Next, we assess whether our attack decreases the naturalness of answers (H5) using a left-tailed Mann-Whitney U test. Here, our collected data is also consistent with H5 ($n_1=$ $n_2=$ 30, $U_{control}=$ 627, $U_{adv}=$ 273, $p=$ 0.004, $Z=$ -2.6174, $r=$ 0.34). This result indicates that human graders perceive student answers with inserted adjectives and adverbs as less natural. We hypothesized that graders would be able to sense the manipulation but not identify it as a cheating attempt (H6).  

Equivalently to our inferiority test conducted on the responses' correctness, we utilize two one-sided Mann-Whitney U tests to test whether our attack increases the mistrust of human graders. We also selected -$\infty$ as lower bound and 0.5 as upper bound. We found that human graders in the experimental group generally thought the students were cheating less or as often as in the control group ($n_1=$ $n_2=$ 30, $U_{control}=$ 576, $U_{adv}=$ 324, $p=$ 0.031). A similar trend can be observed when asking whether graders would take action based on their suspicions. In the control group, graders reported the intention of acting 14 times (N=120). Conversely, graders only wanted to act 5 times (N=90) on the adversarial examples. Graders declined to speak with the student, superior or take disciplinary action for all other answers they rated with at least 4 on the mistrust scale. Examples for the most suspicious responses can be seen in Table \ref{sus}. The examples also illustrate a concerning phenomenon that one of the annotators reported (translated from German):``Generally, I find it difficult to differentiate between bad English and unnatural responses."

\begin{table}
\centering
\caption{Examples of the most suspicious responses from the control (top) and the adversarial group (bottom). The student answer received 2 votes for action, while the adversarial example received 1 vote for action.}\label{sus}
\begin{center}
\begingroup
\setlength{\tabcolsep}{6pt}
\begin{tabularx}{\textwidth}{|l X|} 
\hline
Question: & If Phil, a geologist, wants to test for calcite while in the field, what should he bring with him? (an acid such as vinegar). Describe what Phil should do to test for calcite and what he would observe. \\
& \\
Reference: & Put acid on a rock. If the acid fizzes, Phil would know that the rock has calcite. \\
Answer: & He would put vinegar on the rock get the strange then is quiet it and see if there is calcite. \\
\hline
Question: & The sand and flour in the gray material from mock rocks is separated by mixing with water and allowing the mixture to settle. Explain why the sand and flour separate. \\
& \\
Reference: & The sand particles are larger and settle first. The flour particles are smaller and therefore settle more slowly. \\
Adversarial: & Because one is heavy and \textcolor{red}{high} one is not.  \\
\hline
\end{tabularx}
\endgroup
\end{center}
\end{table}

\subsubsection{Inter-Annotator Agreement}

As discussed in Section \ref{sec::human_eval}, human judgements can be subjective and inconsistent. For this reason, it is common in the NLP field to employ multiple annotators and report their agreement. The inter-annotator agreement provides a measure for how consistent judgements are across annotators. 
Similar to related work, we select Krippendorff's Alpha to estimate our annotators' agreement. As can be seen in Table \ref{alpha}, $\alpha$ is relatively low compared to the broadly applied benchmark of 0.67 \citep{krippendorff2018content}. For the highly subjective and open mistrust question, a low agreement is to be expected. The annotators were informed that some student answers might have been manipulated to fool automatic grading models but not schooled on how such a manipulation could look like. The low agreement ($\alpha=$ 0.13) and slight systematic disagreement ($\alpha=$ -0.11) indicate that the annotators developed individual theories of what cheating would entail in an automatically graded environment.

Additionally, there was a moderate negative Spearman's rank correlation ($\rho$) between mistrust and naturalness ($\rho=$ -0.41) as well as mistrust and correctness ($\rho=$ -0.51) in the control group. In contrast, the correlations in the experimental group were much weaker with $\rho=$ 0.20 and $\rho=$ 0.07, respectively. This indicates that graders suspect poorly written and wrong answers in the absence of other clues.
We will further discuss this behavior and possible ramifications in Section \ref{recs}.

While low inter-annotator agreement is a phenomenon commonly observed in natural language evaluation \citep{amidei-etal-2019-agreement}, we were surprised to see $\alpha$ below 0.3 for naturalness. As recommended by \citet{amidei-etal-2019-agreement}, we calculate $\rho$ for each annotator pair to gain more detailed insight compared to $\alpha$'s holistic score. In the control group, one of the annotators is an outlier with pairwise $\rho$'s of 0.14, 0.07 and -0.02.
The rest of the annotators average a moderate to strong correlation of $\rho$ = 0.57 \citep{akoglu2018user,corder2011nonparametric}. We decided against excluding the outlying annotator from further analysis. Their judgment on the other questions was more in line with the majority, indicating a divergent but potentially valid interpretation of naturalness instead of a systematic disregard for the task. In the experimental group, the average $\rho$ is 0.47.

For correctness, the agreement levels are $\alpha=$ 0.51, $\rho=$ 0.6 in the control group and $\alpha=$ 0.55, $\rho=$ 0.61 in the experimental group. Our observed agreement is expected, considering the generally high inter-grader variability of scores assigned in short answer grading tasks \citep{starch1913reliability}.

\section{Discussion \& Conclusion}
\label{discussion}
In summary, we have introduced an adversarial attack strategy developed explicitly for automatic short answer grading scenarios. It first identifies promising adjectives and adverbs in preparation for employing them during an assessment. Our proposed attack reduces a model's accuracy by 8 to 22 percentage points. Further, we conducted a human expert evaluation to measure our attack's influence on the student answers' correctness, naturalness and suspiciousness. In our experiments, the attack did not significantly increase the correctness or suspiciousness but significantly reduced the perceived naturalness of student responses. Finally, we analyzed the adjective and adverb distribution in the training data and the model's confidence to investigate possible reasons for the model's vulnerability. We found that successful adjectives and adverbs appeared more often in the target class or hardly occurred in the training set. Additionally, adversarial examples tended to elicit a lower confidence score in the model than answers that were not vulnerable to this attack.

The following section offers recommendations for educators looking to employ automatic short answer systems in practice. The recommendations are based on our findings and general knowledge about adversarial attacks. Finally, we will discuss the limitations of our experiments and future work in Section \ref{sec:lims}.

\subsection{Recommendations}
\label{recs}
\textbf{Know thy dataset.} This is especially important as more and more off-the-shelf models become available for various tasks. This development makes it easy to treat machine learning models as black boxes without considering the possible consequences of their training process.
However, a training data analysis can reveal statistical correlations that lead to unreliable prediction features. In our experiments, our attack exploited correlations between adjectives/adverbs and the target class. Beyond our work, non-robust features have been demonstrated for many popular datasets \citep{NEURIPS2019_e2c420d9}. One can also utilize adversarial attacks during training to automatically uncover unreliable features. This is also known as \textbf{adversarial training} and is one of the most promising defenses against adversarial attacks \citep{10.5555/3454287.3454589}, even though it is still limited in its effectiveness since it is typically accompanied by a loss in accuracy on clean data and tends to lack generalizability to novel attack strategies. Moreover, knowledge of potential biases in the dataset can help mitigate discrimination of populations that are not well represented in the data \citep{10.1145/3457607}.

\textbf{Beware of low confidence predictions.} The probabilities assigned to each class can be a valuable indication of whether the prediction is trustworthy. While confidence scores are by no means infallible, they can be a warning sign for when a student's answer should be referred to a manual grader. In our experiments, many of the generated adversarial examples could have been caught this way.

\textbf{School personnel in what to expect.} While automatic grading models are making great strides towards human-like performance on some datasets, we would still recommend employing humans in the grading loop. They can double-check low confidence predictions and perform quality control checks. However, it is vital to educate human control graders on how cheating attempts can look like in the automatic grading age. In our inter-annotator agreement analysis, we observed graders developing individual theories of what made student responses suspicious. Their mistrust would also correlate with how unnatural and incorrect they perceived student answers to be. So, in the absence of other clues or knowledge, the graders in our study would falsely suspect low-performing students and students with poor language skills. We believe that educating human graders on different kinds of attacks and how they express themselves in responses could mitigate such discrimination. In general, any detection method would have to be carefully implemented to avoid disadvantaging minorities not well represented in the data.

\textbf{Balance transparency and exposure of vulnerabilities.} It is crucial that students comprehend their grades. Understanding why a particular grade was given is essential to foster acceptance and enable learning from feedback. Here, making the model's decision process transparent to students is a powerful approach to increase understanding. However, transparency may also reveal exploitable weaknesses, such as unreliable features. Having access to the model's inner workings enables more powerful and efficient adversarial attacks. Therefore, one can argue that keeping grading models secret is sensible. Moreover, one may implement measures that make it harder for adversaries to glean information from querying the model. For example, one can limit the number of times students can receive feedback from the model in a time span to humanly reasonable levels, thus, hindering automatic probing.

\subsection{Limitations \& Future Work}
\label{sec:lims}
Finally, we will point out a few limitations of our experiments and ideas for future work. This paper focused on the effects of one adversarial attack strategy. As the space of possible adversarial manipulations is quite large, it will be exciting to see how well other strategies perform. We then plan to utilize developed attacks in adversarial training to make grading models more robust and explore the models' usability and security in practice. Here, one could also investigate other effects of adversarial attack assessment strategies, such as their impact on responses the student would have answered correctly without the adversarial modification. 
Moreover, we assumed that potential attackers would purposefully aim to fool the model into accepting incorrect responses. It would also be interesting to investigate grading models' robustness to non-malicious writing styles and mistakes, such as common typos or varying verbosity levels. 

Additionally, our experiments could be expanded to other automatic short answer grading architectures. So far, we have explored the attack's effectiveness on transformer-based models on various datasets. While the existence of adversarial vulnerabilities is generally believed to be a result of neural networks exploiting unreliable correlations in the training data instead of being a bug of a particular architecture or hyperparameter setup~\citep{NEURIPS2019_e2c420d9}, we can not rule out that other grading models may be significantly less sensitive to our particular attack. Especially classical machine learning models based on engineered features are likely to require attacks tailored to their feature sets.

Lastly, we mainly see two factors restricting the generalizability of our human evaluation. First, the number of samples annotated was not large enough to reliably detect minor effects. Especially for the mistrust hypothesis, a follow-up study with a larger sample size would have to be conducted to rule out the attack making responses slightly more suspicious. Considering our graders took almost an hour to rate 30 responses, we think more annotators and multiple annotation sessions would make sense. 

Second, all of our graders stem from engineering fields and work at a university. It would be interesting to see whether our observations also hold for other fields and other education institutions. Especially American school teachers may be better at differentiating manipulated answers from poorly written ones. While our annotators were accustomed to grading English short answers in their daily lives and speak English proficiently, they were not native speakers. Moreover, they stem from various countries, such as India and Slovenia, and may speak different English dialects. This probably impacted the evaluation of naturalness, as indicated by the low inter-annotator agreement, but we expect only a minor effect on the correctness and mistrust scales.

\bibliographystyle{spbasic}      
\bibliography{ijaied2021}   

\end{document}